\documentclass{article}
\usepackage{spconf,amsmath,graphicx}
\usepackage{subcaption}
\usepackage{amsfonts} 

\usepackage{xcolor}
\usepackage{multirow,epstopdf}
\usepackage{booktabs}
\usepackage{algorithm2e}
\usepackage{amsmath}
\usepackage{bbm}
\usepackage{enumerate}
\usepackage{hyperref}
\usepackage{url}

\title{$\text{H}_{\text{eval}}$: A new hybrid evaluation metric for automatic speech recognition tasks}
                \name{Zitha Sasindran, Harsha Yelchuri, T. V. Prabhakar, Supreeth Rao}
                \address{Department of Electronic Systems Engineering \\
                        Indian Institute of Science \\
                         Bengaluru, India, 560012\\
                         Email: \{zithas, harshay, tvprabs\}@iisc.ac.in, raosupreeth00@gmail.com}

\copyrightnotice{979-8-3503-0689-7/23/\$31.00~\copyright2023 IEEE}

\begin{document}
%
\maketitle
\begin{abstract}

 Many studies have examined the shortcomings of word error rate (WER) as an evaluation metric for automatic speech recognition (ASR) systems. Since WER considers only literal word-level correctness, new evaluation metrics based on semantic similarity such as semantic distance (SD) and BERTScore have been developed. However, we found that these metrics have their own limitations, such as a tendency to overly prioritise keywords. We propose $\text{H}_{\text{eval}}$, a new hybrid evaluation metric for ASR systems that considers both semantic correctness and error rate and performs significantly well in scenarios where WER and SD perform poorly. Due to lighter computation compared to BERTScore, it offers 49 times reduction in metric computation time. Furthermore, we show that $\text{H}_{\text{eval}}$ correlates strongly with downstream NLP tasks. Also, to reduce the metric calculation time, we built multiple fast and lightweight models using distillation techniques without significant reduction in performance.
\end{abstract}
\section{Introduction}
\label{sec:intro}
Word error rate (WER) is the de-facto evaluation metric used by the speech community for assessing the performance of automatic speech recognition (ASR) models. WER is calculated from a literal word-level comparison of ground-truth with the hypothesis. 
Consider the ground-truth `Morale is raised', and the two hypotheses from  ASR be `Moral is raised' and `Morale is razed'. WER is same for both the sentences, but we can clearly see that the former hypothesis makes more sense than the latter which completely negates the meaning of the ground-truth.
Even though WER is flawed and is an imperfect metric for evaluation, several works tend to use WER as an evaluation metric because of the following reasons : (1) existing speech recognition models are benchmarked with WER as the metric, (2) an improved alternative metric does not exist to replace WER.
There is an increasing interest among the speech community to consider an alternate evaluation metric\cite{wernotgood,wer_alternative1}, since WER does not capture the semantic correctness of the hypothesis generated.  



Over the years, many works have attempted to address the flaws of WER by considering weighted WER or by adopting information retrieval approaches as mentioned in \cite{inf_retrieval,weighted_wer1,weighted_wer2}. But all of these metrics are based on word level comparison, and have not considered the semantic-level correctness of the sentences. Meanwhile many prior works based on the pre-trained transformer \cite{transformers} based language models, such as BERT \cite{bert}, RoBERTa \cite{roberta} showed promising results in various downstream tasks such as question answering, and sentiment analysis. Recently, \cite{semanticdistance,semanticdistance2} proposed an alternative metric based on semantic distance (SD) for ASR tasks based on semantic correctness using Siamese-BERT \cite{sbert} models.  Our extensive analysis of this metric on ASR hypotheses shows that using a metric just on semantic correctness is also not an optimal evaluation metric. 
In \cite{bertscore_asr}, authors proposed to utilise BERTScore \cite{bertscore} for assessing ASR model quality on disordered speech. Our thorough analysis shows that $\text{H}_{\text{eval}}$ works similarly to BERTScore while being faster.

In this work, we contrast our proposed evaluation metric to various scenarios where other metrics fall short. We first focus on building a method to get the SD score as explained in \cite{semanticdistance}. We observe that this measure is biased heavily towards the words which contributes more to the meaning of the sentence (keywords), and sometimes fails to give a proper weightage to the remaining words. Hence, we propose a new weighted evaluation metric $\text{H}_{\text{eval}}$, which is a weighted combination of SD score and non-keyword error rate. We also explain the methodology to extract the keywords from the ground-truth, which is crucial for generating our metric. Further, we use the knowledge distillation \cite{knowledge_distillation}  approach to create lighter and faster versions of these siamese-BERT models, to obtain the SD score for faster and quick inferences without much compromise in accuracy.  


This paper is organised as follows. 
In Section \ref{sec:hybridmetric}, we discuss the proposed hybrid evaluation metric. In Section \ref{sec:experimental results}, we explain the experimental setting and the results and finally, concluding remarks are presented in Section \ref{sec:conclusion}.

\section{A hybrid evaluation metric}
\label{sec:hybridmetric}

In this section, we first discuss the methodology for generating the SD score, and BERTScore, followed by the limitations of them, and then propose our new hybrid evaluation metric.


\noindent{\textbf{SD and BERTScore}: }Siamese models \cite{sbert} outperform other models in semantic textual similarity (STS) tasks, in which two models connected by a loss function are fed different inputs. We employ the RoBERTa model in the siamese configuration (SRoBERTa), followed by a mean pooling of the sentence embedding tensors to derive fixed size vectors for sentence pairs. To calculate the semantic distance, we use a sub-word tokenizer during the pre-processing stage, that learns to tokenize a corpus into `word pieces', starting from an initial alphabet containing all the characters in the corpus. These word pieces allow the models to accurately score similar sentences that share same tokens. Further, we use cosine similarity to compute the similarity between the two fixed sized embedding vectors generated by both the models. The SD \cite{semanticdistance} score between two sentences $S_1$ and $S_2$ is:
\begin{align}
SD~(S_1,S_2) = 1 - \frac{e_{S_1}~.~ e_{S_2}}{||e_{S_1}||~ ||e_{S_2}||}
\end{align}
where $e_{S_i}$ denotes the respective embedding vector generated from our model for a sentence $S_i$. SD is confined between 0 and 1, where lower scores indicate stronger semantic similarity, and vice versa. When subwords are present in the sentence, the SD score improves because of the subword tokenizer.
For example, GT denotes the ground-truth and H1 and H2 are two hypotheses from the ASR model, i.e., GT = `Smoking', H1 = `Smoke', and H2 = `Something'. Then $\text{SD}(\text{GT},\text{H1} )$ = 0.09, which is less than $\text{SD}(\text{GT},\text{H2} )$ = 0.85.

We begin with an example by demonstrating the limitations of SD. Consider GT as `This is your captain speaking', and let H1 be `des s ur captain speaking' and H2 be `This is your kepten speaking'. Then $\text{SD}(\text{GT},\text{H1})$ = 0.29 and $\text{SD}(\text{GT},\text{H2})$ = 0.41. We can see that `captain' and `speaking' are the keywords of the sentence.
H1 has all the keywords correct, but all the non-keywords are incorrect.  In case of H2, all the words except `captain' are correct. Although it is natural to select H2 as the best hypothesis as it is more interpretable and meaningful, SD of H1 lower than H2. 
Despite the fact that H2 has fewer errors, SD tends to favour H1 merely because keywords are correctly predicted. For instance, the mere presence of the keyword `captain' in H1 is neglecting the errors in the other non-keywords. The SD score is biased towards keywords that appear to add meaning to the sentence and sometimes fails to provide the remaining words an appropriate weightage. Thus, for ASR tasks, SD appears to be limited as a metric as it is specific to tasks where keywords are important. 

The process of calculating BERTScore \cite{bertscore} involves first obtaining contextual embeddings for each word token, followed by computing pairwise cosine similarities between the token embeddings of the GT and hypothesis. The maximum similarity is then selected. However, BERTScore is computationally expensive and relies on a pre-trained BERT model, which can encounter challenges with out-of-vocabulary words. Additionally, as mentioned by the authors in \cite{bertscore_asr}, the objective is to combine the WER and a semantic distance based score to better fit the data than using either parameter alone.
In contrast to \cite{semanticdistance} and \cite{bertscore_asr}, our proposal is to arrive at a metric that considers both semantic correctness and the number of word errors. 


\noindent{\textbf{Our approach}: }To get a quick overview of a text content, it can be helpful to extract keywords that concisely reflect its semantic context.
We aim to find the keywords and non-keywords and give appropriate weightage to the words unlike SD.
Our intent is to extract the keywords using the same model used for generating SD score, instead of using a separate model like KeyBERT \cite{keybert} whose size is of 135 MB, thereby saving the memory. Our approach of keyword extraction is similar to that of KeyBERT, where, we compute the cosine similarity between the word and sentence embedding.
Our keyword extraction approach is as follows:  
\begin{align*}
S&=\{x:x = SD~(GT,w_i),~\forall~ w_i \in~ W\}\ \\
K&=\{w_i:MinMax~(S~[w_i])~<~\gamma,~\forall~ w_i \in~ W \} \\
NK&=W-K
\end{align*}
where $W$ is the set of words in GT. We calculate the SD score for GT and each word in set $W$, and form the set $S$. Then, we normalise the scores in $S$ by using Min-Max scaling, and construct the  set $K$ by thresholding using $\gamma$ parameter. Next, the set $NK$ is obtained by removing words of $K$ from $W$. Thus, we create two sets of words from GT namely: $K$ and $NK$ containing keywords and non-keywords respectively.

Let $N_{nk}$ denote the number of non-keywords present in GT, i.e., $N_{nk} = |NK|$. $N_{wk}$ and $N_{wnk}$ are the number of wrongly recognised keywords and non-keywords respectively in the hypothesis. To compute $N_{wk}$ and $N_{wnk}$ from hypothesis, we use an edit-distance based alignment between GT and hypothesis.  Finally, we calculate the non-keyword error rate (NKER) as the ratio of $N_{wnk}$ to $N_{nk}$.
 We propose a hybrid evaluation metric which is a weighted combination of SD and NKER, and is calculated as follows:
 \begin{align}
 H_{eval} &= \alpha_1*SD~+~\alpha_2*NKER \\
 \alpha_1 &=\frac{N_{wk}*p}{N},~~\alpha_2=\frac{N_{wnk}}{N}
  \end{align}
where the first term is used to monitor semantic correctness by prioritising keywords, and the second term monitors the non-keywords. 
We calculate the weighting coefficients $\alpha_1$ and $\alpha_2$  for SD and NKER based on the number of wrong keywords and wrong non-keywords respectively, present in the hypothesis.  We define $p$ as the ratio of and $N$ and $N_{k}$  which determines the weight one keyword has with respect to other words in the sentence.  Thus, $p$ strikes a balance between the importance given to keywords and the non-keywords.

The main motivation behind the formulation is to arrive at a metric which is not biased towards keywords.
But first, we need to find out what are the keywords in the given sentence. Once, keywords are determined, we monitor how much percentage of keywords and non-keywords are decoded wrongly in the hypothesis and use these parameters as weighing coefficients for SD and NKER. One important parameter used in weighing is `p' which is used to control the amount of importance given to keywords over non-keywords. If all the keywords are decoded correctly, there is no need to monitor keywords anymore. In such case $a_1$ gets value of 0, making the whole equation focus on non keywords. Similarly if all the non-keywords are decoded correctly, $a_2$ gets value of 0, making the whole equation focus on keywords.

  \begin{table}[!h]
	\centering
	\caption{Comparison of WER with SD and  $\text{H}_{\text{eval}}$}
    \resizebox{0.46\textwidth}{!}{%

	\begin{tabular}[t]{llccc}
		\toprule
		&\textbf{~~~~~~~~~~~~~Examples}&\textbf{WER}&\textbf{SD}&\textbf{$\text{H}_{\text{eval}}$}\\
		\midrule
		1&\textbf{GT : The flight is about to land}&&\\
		&H1 : The fite is about to lamt&0.33&0.72&0.72\\ 
		&H2 : Te flight s about to land&0.33&\textbf{0.11}&\textbf{0.17}\\
		\midrule
		2&\textbf{GT : Data set needs to be cleaned }\\
		&H1 : Dase needs to be cleaned&0.33&0.37&0.18\\ 
		&H2 : Dataset needs to be cleaned&0.33&\textbf{0.23}&\textbf{0.11}\\
		\midrule
		
		3&\textbf{GT : Whomsoever it is concerned}&&\\
		&H1 : hm so er it is concerned&0.75&0.34&0.17\\
		&H2 : whom so ever it is concerned&0.75&\textbf{0.09}&\textbf{0.05}\\
		
		\bottomrule
	\end{tabular}
    }
	\label{tab:WER_vs_SD}
\end{table}

\begin{table}[!h]
	\centering
	\caption{Comparison of $\text{H}_{\text{eval}}$ with SD, and BERTScore ($\text{F}_{\text{BERT}}$). 
    Lower values of SD and $\text{H}_{\text{eval}}$ are considered better, while higher values are preferable for the $\text{F}_{\text{BERT}}$.}
    \resizebox{0.46\textwidth}{!}{%
 
	\begin{tabular}[t]{llccc}
		\toprule
		&\textbf{~~~~~~~~~~~~~Examples}&\textbf{SD}&\textbf{$\text{H}_{\text{eval}}$}
        &\textbf{$\text{F}_{\text{BERT}}$}\\
		\midrule
		1&\textbf{GT : I am in love with the nature}\\ 
        &\textbf{in Ontario}&&\\
		&H1 : m n love vid de nature\\ &n Ontario&\textbf{0.12}&0.67&0.85\\ 
		&H2 : I am in love with the netuore\\ &in Ontario&0.39&\textbf{0.13}&\textbf{0.95}\\
		\midrule
  2&\textbf{GT : I am too shy to speak}\\ &\textbf{to that many people.}&&\\
		&H1 : Im to shy tu speak\\ &two dat many people.&\textbf{0.28}&0.51&0.87\\ 
		&H2 :I am too shy to speak\\ &to that many ppal.&0.31&\textbf{0.10}&\textbf{0.96}\\
		\midrule
		  3&\textbf{GT : He had a ridiculous idea}\\ &\textbf{and I was very upset by it.}&&\\
		&H1 : Hi hada ridiculous idea\\ &nd ivas wery upset buy itt.&\textbf{0.44}&0.75&0.86\\ 
		&H2 : He had a ridikulas idea\\ &and I was very up set by it.&0.61&\textbf{0.41}&\textbf{0.92}\\
		\bottomrule
	\end{tabular}
\label{tab:SD_vs_Hsd}
}
\end{table}

\section{EXPERIMENTAL RESULTS}
\label{sec:experimental results}
We conducted extensive experiments to investigate the efficacy of our proposed $\text{H}_{\text{eval}}$, and the results are presented in this section. Throughout our experiments, the parameter $\gamma$ was set to 0.4. The range of WER, $\text{F}_{\text{BERT}}$, and $\text{H}_{\text{eval}}$ are $[0,100],[0,1]$, and $[0,100]$ respectively. All the datasets and the distilled models employed are made available\footnote{\url{https://drive.google.com/drive/folders/1wy8NixLbh6bwyS-suedpsHqNs0I-fgBl?usp=drive_link}}.
\subsection{Comparison between WER, SD, $\text{F}_{\text{BERT}}$  and $\text{H}_{\text{eval}}$}

\label{sec:sd-and-wer}
In this section, we examine how WER, SD and $\text{H}_{\text{eval}}$ values differ from one another with several examples, and discuss the benefits of $\text{H}_{\text{eval}}$ over WER and SD.




We will take some examples to demonstrate the need for a new evaluation metric which improves upon WER. From Table \ref{tab:WER_vs_SD}, consider the two hypothesis H1 and H2 generated from ASR models. With WER, although the number of errors are the same, in the first example, WER fails to track the improvement of `fite' and `lamt' to `flight' and `land', due to the deterioration of `is' and `the'  to `s' and `te' at the same time. Since `flight' and `land'  offer more contextual information than `is' and `the', SD and $\text{H}_{\text{eval}}$ decreases  significantly for H2.
In the second and third example, H2 has a lower SD and $\text{H}_{\text{eval}}$ value because of the sub-word tokenizer's ability to capture subwords. 

While the results in Table \ref{tab:WER_vs_SD} are encouraging, Table \ref{tab:SD_vs_Hsd} shows that SD is unreliable as an evaluation metric. In Table \ref{tab:SD_vs_Hsd}, if we consider the first example, even though H1 has six errors when compared to H2's one error, SD still says H1 is a better sentence because of the word `nature' in H1. The same case can be observed in examples 2 and 3, where SD is biased towards sentences with more number of errors just because that sentence has a few correct keywords \{shy, speak, people\} from H1 of example 2 and \{ridiculous, upset\} from H1 of example 3.
We can clearly infer that SD is excessively prioritising sentences with correctly predicted keywords while ignoring all other words.

In summary, while Table \ref{tab:WER_vs_SD} shows the limitations of WER, Table \ref{tab:SD_vs_Hsd} demonstrates that BERTScore and our metric $\text{H}_{\text{eval}}$ works similarly in the cases where SD fails.
However, the key takeaway to note is that in both the tables, $\text{H}_{\text{eval}}$ evaluates the sentences closer to the ground-truth since it considers both semantic correctness and non-keyword error rate, thus outperforming SD and WER.



\begin{table*}[!t]
\centering
\caption{Comparison of $\text{H}_{\text{eval}}$ with Intent recognition and Named-Entity recognition metrics}
\begin{tabular}{cccccc}
\toprule
     \textbf{Datasets} & \textbf{WER} & \textbf{$\text{H}_{\text{eval}}$} & \textbf{Intent Recognition Accuracy} & \textbf{Named-Entity Error}\\
     &\textbf{(\%)}&&\textbf{(\%)}& \textbf{Rate (\%)}\\
     \midrule
         Set A (ASR Outputs) & 27 & 0.13  & 88.16 & 49.28\\ 
         Set B & 27 & 0.02    & 90.62 & 21.06 \\
         Set C & 27 & 0.30    & 81.25 & 58.91 \\
    \midrule
\end{tabular}%
\vspace{-7pt}

\label{tab:intent_ner}
\end{table*}

\begin{figure*}[!t]
     \centering
     \begin{subfigure}[b]{0.3\textwidth}
         \centering
         \includegraphics[width=\textwidth]{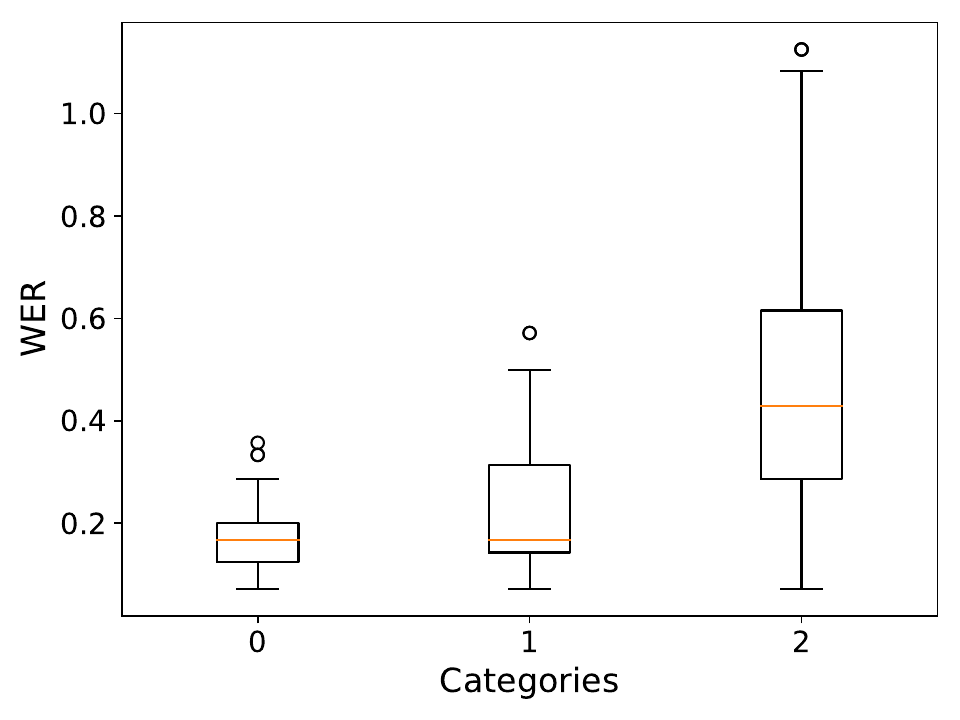}
         \caption{WER}
         \label{fig:WER}
     \end{subfigure}
     \hfill
     \begin{subfigure}[b]{0.3\textwidth}
         \centering
        \includegraphics[width=\textwidth]{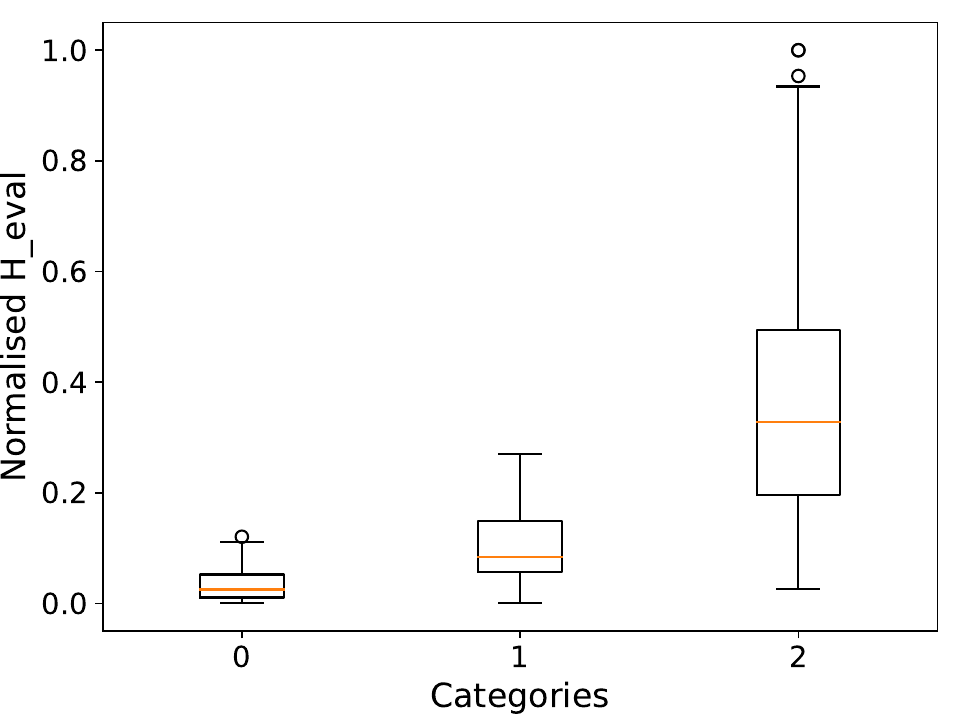}
         \caption{$\text{H}_{\text{eval}}$}
         \label{fig:h_eval}
     \end{subfigure}
     \hfill
     \begin{subfigure}[b]{0.3\textwidth}
         \centering
         \includegraphics[width=\textwidth]{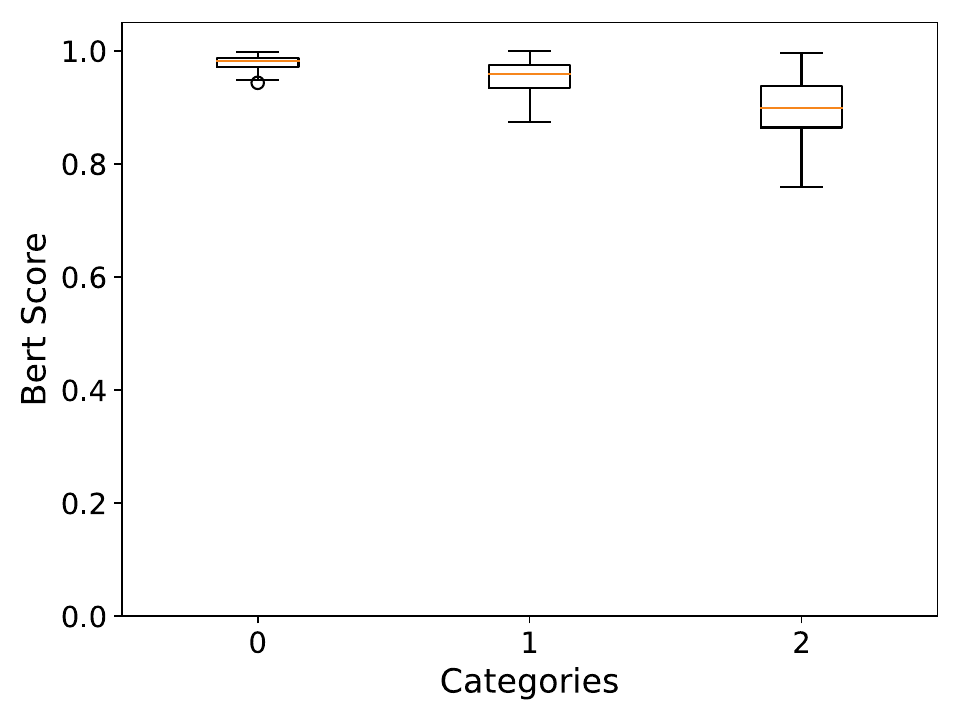}
         \caption{BERTScore}
         \label{fig:bert}
     \end{subfigure}

        \caption{Distribution of various metrics by error assessment.}
        \label{fig:disordered}

\end{figure*}

\subsection{ASR Analysis}
\label{asr}
We use DeepSpeech2 \cite{ds2}, Whisper\cite{radford2022robust}, Wave2vec2\cite{baevski2020wav2vec} models to  demonstrate the effectiveness of $\text{H}_{\text{eval}}$ and other evaluation metrics for realistic ASR systems. In order to show the effectiveness of $\text{H}_{\text{eval}}$, we conducted two different experiments. 


In the first experiment, we show that $\text{H}_{\text{eval}}$ correlates with human understanding. To accomplish this, we took Torgo data set \cite{torgo} which contains speech from speakers with dysarthria. We created inference for this data set using the above-mentioned models. Each model inferred 200 samples of the data set. Next, we asked experts to rank how close the inferred sentence is to the ground truth. The criteria to rank is very similar to that used in BERTScore.

In the second experiment,  we co-evaluate $\text{H}_{\text{eval}}$ with other natural language understanding-based metrics such as intent recognition accuracy and named-entity error rate and show that $\text{H}_{\text{eval}}$ is highly correlated with these tasks.
For this purpose, we create a speech corpus of 500 samples from ATIS \cite{atis} test dataset by filtering out shorter sentences. The speech samples are generated using a text-to-speech \cite{speechelo} system with five different speakers.  In order to evaluate our $\text{H}_{\text{eval}}$ metric, we created a hypotheses set (Set A) by feeding these samples through the Deepspeech2 model. To expand our evaluation, we create two additional hypotheses sets, Set B and Set C, that have the same number of errors (same WER) as the baseline ASR-generated hypotheses set (Set A). Set B is designed to have a better $\text{H}_{\text{eval}}$ score than Set A. It is created by including errors that minimally affect the meaning of a sentence, such as inserting articles or splitting words in the corresponding reference GT. On the other hand, Set C is designed to have a worse $\text{H}_{\text{eval}}$ score than Set A. It is generated by introducing errors such as either removing or replacing keywords with random words.

\subsubsection{Disordered speech}
Fig. \ref{fig:disordered} presents the results obtained with the first experiment. This experiment utilises sentences grouped into 3 categories, Category 0 has sentences in which the meaning is completely captured. Category 1 has sentences in which meaning is mostly captured (some errors). Category 2 has sentences in which the meaning is completely different. To ensure a more comparable analysis, we applied a normalisation technique to the $\text{H}_{\text{eval}}$ value, scaling it to a range between 0 and 1. 


Fig. \ref{fig:WER} shows that even if WER rises when we move from category 0 to category 2, its standard deviation is higher than that of $\text{H}_{\text{eval}}$ as shown in Fig. \ref{fig:h_eval}. In case of BERTScore as shown in Fig. \ref{fig:bert}, its standard deviation is less than that of $\text{H}_{\text{eval}}$. However, its performance on category 2 (containing samples with a WER $>$ 1) is unexpected as it assigns them a higher value of 0.75 instead of 0. This is despite the sentences having different meaning. This suggests that BERTScore may not fully capture the discrepancy between the generated and reference sentences in terms of semantic similarity. Whereas, our $\text{H}_{\text{eval}}$ metric assigns a higher value indicating that the sentences are totally different.
Overall $\text{H}_{\text{eval}}$ performs well.

We also observed that on an average of sentences with 10 words, the metric calculation time taken by BERTScore is around 1.95 seconds while $\text{H}_{\text{eval}}$ takes 0.04 seconds. This indicates that BERTScore generally requires more computational time compared to  $\text{H}_{\text{eval}}$ for evaluating sentences of similar length.
\subsubsection{Intent recognition and NER}
We utilise models from Spacy \cite{spacy}, an open-source software library, to extract intents and named-entities from the ASR hypotheses. The models employed for this purpose are trained on the ATIS \cite{atis} training set, comprising of 4,978 sentences with 17 distinct intent categories, and 79 named-entity categories, respectively. 

\begin{table*}[!t]
\caption{Spearman rank correlation between the cosine similarity of sentence embeddings and the gold labels for various STS tasks using teacher and our student models. All models use the siamese network architecture.}
\vspace{-3pt}

\setlength{\tabcolsep}{2pt}
\resizebox{\textwidth}{!}{%

\begin{tabular}{l|cccccc|cccccccc}

    \toprule
    \textbf{Model }           &\textbf{\# Layers }    & \textbf{\# Hidden} &\textbf{\# Attention} &\textbf{\# Parameters} & \textbf{Size} & \textbf{Latency} &\textbf{STS-12} &\textbf{STS-13} &\textbf{STS-14} &\textbf{STS-15} &\textbf{STS-16} &\textbf{SICK-R} &\textbf{STS-B}  &\textbf{Average}  \\
    &  &\textbf{Dim} &\textbf{Heads} &\textbf{ (M)}&\textbf{ (MB)} &\textbf{(ms/} & & & & & & & &\\
    & & & & & &\textbf{sample)} &  & & & & & & & \\
    
    \midrule
    SRoBERTa          &12    &768     &12   &124.6 &499.0  & 129.32  &0.787  &0.901  &0.882  &0.9    &0.848  &0.797  &0.871  &0.855  \\
    \midrule
    SRoBERTa [1,4,7,10]&4     &768     &12   &67.94 &271.8& 41.12   &0.751  &0.886  &0.86   &0.892  &0.836  &0.793  &0.867  &0.840 \\
    STinyBERT         &4     &312     &12   &14.35 &57.4 & 6.37   &0.793  &0.862  &0.834  &0.88   &0.804  &0.784  &0.824  &0.825   \\
    SNanoBERT     &4    &128 &4   & 4.78&19.2  & 3.33&0.732  &0.851  &0.807&0.866&0.801 &0.761  &0.81 &0.804 \\
    SPicoBERT         &1     &64      &1   &2.07 &8.3   & 0.65  &0.772  &0.835  &0.783  &0.849  &0.771  &0.756  &0.778  &0.792  \\
    \bottomrule
    \end{tabular}%
}
\vspace{-6pt}
\label{tab:benchmark}
\end{table*}



The objective is to explore the relationship between our $\text{H}_{\text{eval}}$ metric and intent recognition accuracy on ASR hypotheses sets. We extracted intents from Set A, Set B, and Set C, and calculated their intent recognition accuracy. As depicted in Table \ref{tab:intent_ner}, Set C showed the highest $\text{H}_{\text{eval}}$ metric and the lowest intent accuracy, primarily due to incorrectly identified keywords in the hypothesis. Set A had a slightly lower $\text{H}_{\text{eval}}$ value but displayed better intent accuracy, whereas Set B had the lowest $\text{H}_{\text{eval}}$ value but reported the highest intent accuracy. Despite the sets having the same WER, we observed that as $\text{H}_{\text{eval}}$ decreased, intent accuracy increased. Hence, we can conclude that the $\text{H}_{\text{eval}}$ metric evaluates ASR hypotheses in a manner that is similar to intent recognition accuracy.


Subsequently,  we investigate the correlation between the $\text{H}_{\text{eval}}$ metric and the named-entity error rate on ASR hypotheses sets. As presented in Table \ref{tab:intent_ner}, Set C, which has the highest $\text{H}_{\text{eval}}$, also has the highest entity error rate due to misspelt keywords. In contrast, Set B, with the lowest $\text{H}_{\text{eval}}$, has the lowest entity error rate since non-keywords are misspelt. We can conclude that the entity error rate is directly proportional to the $\text{H}_{\text{eval}}$ value. Hence, our proposed metric aligns well with the named entity error rate while evaluating ASR hypotheses. 

In summary, we can infer that $\text{H}_{\text{eval}}$ can serve as a comprehensive and effective evaluation metric not only for ASR tasks but also for intent recognition and NER tasks.

\subsection{Motivation for lighter and faster language models}
Given the sophistication of language model based metrics as compared to WER, it is imperative to explore lighter and faster models.   

Although the SRoBERTa model is more accurate, its large size (500MB from Table \ref{tab:benchmark})  makes it significantly slow for inference. From Table \ref{tab:benchmark}, we require an average of 129.32 ms for metric calculation.
Training ASR models on edge devices in real-world environments to improve their practicability for on-device voice personalisation and privacy preserving applications (e.g. federated learning) is becoming more popular. 
Evaluation metrics such as $\text{H}_{\text{eval}}$  also play a critical role in devising a better stopping criteria while training the model. This is important as the weights can get corrupted by indiscriminately continuing training, leading to overfitting the model.   
Using such large language models just for the sake of stopping criteria is not ideal in these scenarios.
Therefore it is important to scale down the model to lighter and faster versions to comply with the deployment constraints.


 


\subsubsection{Creating lighter and faster models}
We use a SRoBERTa \cite{sbert} model fine tuned on the STSB \cite{stsb} dataset as the teacher model, and our custom models with lesser layers (transformer blocks), lesser hidden dimensions and lesser attention heads as the student models. We carry out distillation with the task of generating embeddings using the AllNLI \cite{mnli,snli} and English Wikipedia datasets and use the STSB to evaluate our student models. As the student models have fewer hidden dimensions, we perform dimensionality reduction using principal component analysis.
We train our models using mean squared error loss function which is calculated between the embeddings of the teacher model and the student models. Our models are distilled at a learning rate of 1e-4.

\subsubsection{Results}

We benchmark all the distilled models as well as the teacher model (SRoBERTa) on a collection of sentence similarity benchmark datasets such as STS-12\cite{sts2012}, STS-13\cite{sts2013}, STS-14\cite{sts2014}, STS-15\cite{sts2015}, STS-16\cite{sts2016}, SICK-R\cite{sickr}, STS-B\cite{stsb} and have tabulated the spearman correlation coefficient between the cosine-similarity of the sentence embeddings and the gold labels as shown in Table \ref{tab:benchmark}. We created four distilled siamese models SRoBERTa[1,4,7,10], STinyBERT, SNanoBERT, and SPicoBERT whose sizes ranges from 271.8 MB to 8.3MB. 
The smallest model SPicoBERT achieves an average correlation coefficient of 0.792, which is only 0.063 lower than SRoBERTa while being 60.1x smaller and 198.3x faster than SRoBERTa.

\section{SUMMARY AND CONCLUSIONS}
\label{sec:conclusion}

In this work, motivated by the shortcomings of the existing evaluation metrics, we propose $\text{H}_{\text{eval}}$, a new hybrid evaluation metric for ASR systems that takes into account both semantic correctness and non-keyword error rate. Our extensive evaluation shows that $\text{H}_{\text{eval}}$ performs significantly well in scenarios where WER and SD perform poorly. The metric computation time is 49 times faster compared to BERTScore. This might have implications on the overall training time when used for stopping criteria. Furthermore, we demonstrate that $\text{H}_{\text{eval}}$ exhibits a strong correlation with human assessment and metrics such as intent recognition accuracy and named-entity error rate. We also built multiple fast and lightweight models using distillation to generate SD score. Hence, $\text{H}_{\text{eval}}$ takes a holistic approach as an evaluation metric.

\bibliographystyle{IEEEtran}
\bibliography{mybib}

\end{document}